\documentclass[runningheads]{llncs}
\usepackage[T1]{fontenc}
\usepackage{graphicx}
\usepackage{booktabs}
\usepackage{multirow}
\usepackage{hyperref}

\begin{document}
\title{How Annotation Trains Annotators: Competence Development in Social Influence Recognition}
\titlerunning{Competence Development in Social Influence Recognition}

\author{
Maciej Markiewicz\inst{1}\orcidID{0009-0004-2882-6741} \and
Beata Bajcar\inst{1}\orcidID{0000-0001-5044-4070} \and
Wiktoria Mieleszczenko-Kowszewicz\inst{1}\orcidID{0000-0002-3948-268X} \and
Aleksander Szczęsny\inst{1}\orcidID{0009-0003-6808-2321} \and
Tomasz Adamczyk\inst{1}\orcidID{0009-0005-9703-4630} \and
Grzegorz Chodak\inst{1}\orcidID{0000-0002-9604-482X} \and
Karolina Ostrowska\inst{2}\orcidID{0009-0004-9959-2487} \and
Aleksandra Sawczuk\inst{2}\orcidID{0009-0002-0677-7905} \and
Jolanta Babiak\inst{1}\orcidID{0000-0002-6604-7763} \and
Jagoda Szklarczyk\inst{3}\orcidID{0009-0004-6784-8273} \and
Przemysław Kazienko\inst{1}\orcidID{0000-0001-5868-356X}
}

\institute{
Wrocław University of Science and Technology, Wrocław, Poland \\ \email{\{name.surname\}@pwr.edu.pl} \and
University of Silesia in Katowice, Katowice, Poland \and
SWPS University, Wrocław, Poland
}

\authorrunning{M. Markiewicz et al.}
%
%
\maketitle              
\begin{abstract}
Human data annotation, especially when involving experts, is often treated as an objective reference. However, many annotation tasks are inherently subjective, and annotators' judgments may evolve over time. This study investigates changes in the quality of annotators' work from a competence perspective during a process of social influence recognition. The study involved 25 annotators from five different groups, including both experts and non-experts, who annotated a dataset of 1,021 dialogues with 20 social influence techniques, along with intentions, reactions, and consequences. An initial subset of 150 texts was annotated twice -- before and after the main annotation process -- to enable comparison. To measure competence shifts, we combined qualitative and quantitative analyses of the annotated data, semi-structured interviews with annotators, self-assessment surveys, and Large Language Model training and evaluation on the comparison dataset. The results indicate a significant increase in annotators' self-perceived competence and confidence. Moreover, observed changes in data quality suggest that the annotation process may enhance annotator competence and that this effect is more pronounced in expert groups. The observed shifts in annotator competence have a visible impact on the performance of LLMs trained on their annotated data.

\end{abstract}

\noindent\textit{\textbf{Preprint:} Accepted to AIED 2026: The 27th International Conference on Artificial Intelligence in Education. This preprint has not undergone peer review or any post-submission improvements or corrections. The Version of Record of this contribution is published in LNCS vol 16584, and is available online at
\url{https://doi.org/10.1007/978-3-032-29763-1_36}. Please cite the published version.}
\section{Introduction}\label{introduction}

Human annotation is foundational for creating high quality datasets and has been extensively studied. Expert annotation is often treated as objective ground truth, and aggregating multiple annotators typically yields satisfactory results despite individual errors. However, in subjective or partially-subjective tasks, such as the recognition of social influence, or when annotators are not experts, even human annotation may not be accurate or may change over the course of the process \cite{fleisig-etal-2024-perspectivist}. The aim of this study is to identify the changes in annotators' work and analyze their origins; whether it is the effect of increased competence, random noise, or a varying worldview. We then want to assess the impact of these changes on data quality and AI model training. The study is conducted in conjunction with our unpublished social influence recognition dataset but focuses solely on the annotation process. We formulate the following research questions:

\begin{description}
\item[\textbf{RQ1}] How does the competence of annotators regarding social influence recognition change over the course of the annotation process?
\item[\textbf{RQ2}] How does the quality of the annotators' work change throughout the annotation process?
\item[\textbf{RQ3}] Does using data from the beginning or the end of the annotation process for AI model training lead to changes in model performance?
\end{description}

\section{Related Work}\label{related}

Recent work on subjective tasks, such as the recognition of social influence, challenges the ground truth assumption, proposing perspectivist approaches that preserve individual annotator viewpoints rather than aggregating them \cite{fleisig-etal-2024-perspectivist,mokhberian-etal-2024-capturing}. To evaluate the reliability of these viewpoints, \cite{abercrombie-etal-2025-consistency} emphasize using intra-annotator agreement to distinguish valid subjective interpretations from noise, rather than relying solely on inter-annotator agreement, which has been found to be a weak quality measure in subjective tasks \cite{bassi-etal-2025-annotating}. However, annotator performance is rarely stable. \cite{abercrombie-etal-2023-temporal} observe that intra-annotator agreement diminishes over time, while \cite{bassi-etal-2025-annotating} show that annotators' behavior has evolved significantly under supervision.

The observed evolution in annotator behavior can be framed through learning science. \cite{Renkl2014} argue that example-based learning is a reliable method of knowledge acquisition, similar to learning from one's own experience \cite{kolb1984}. Similarly, crowd-sourced annotation tasks can be treated as learning environments where workers actively gain knowledge \cite{10.1145/2858036.2858268,Hata2017}. This aligns with the definition of competence as an integration of knowledge and skills that develop over time \cite{vitello_greatorex_shaw_2021}. Empirical studies demonstrate that non-experts, such as language learners, can contribute high-quality data as their skills improve \cite{yoo-etal-2023-rethinking}, while \cite{10.1162/coli_a_00436} show that annotation curricula allow annotators to implicitly learn task schemes. Our study extends this by investigating how this natural competence acquisition affects the utility of the resulting data for model training.

\section{The annotation process}\label{process}

\begin{figure}[htbp]
    \centering
    \includegraphics[width=1\linewidth]{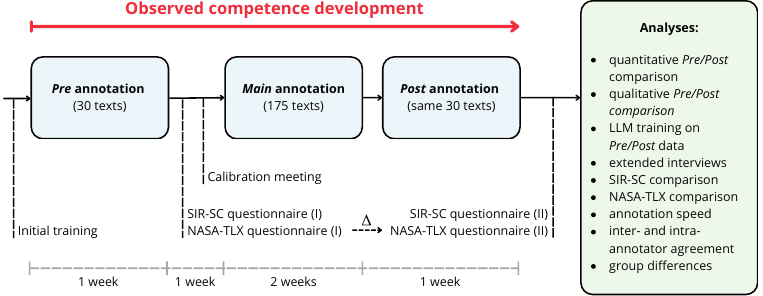}
    \caption{Overview of the studied annotation process with competence shift analyses.}
    \label{fig:overview}
\end{figure}

The studied annotation process aimed to create a dataset of 1,021 AI-generated dialogues depicting social influence targeted at adolescents, Figure~\ref{fig:overview}. Each text has been verified and adjusted to sound natural, both by experts and by a single super-annotator. Each dialogue was annotated by 5 annotators, one from each group. The groups included experts: psychologists and communication experts, and non-experts: adolescents, parents, and teachers. The annotators' task was to label texts with the following: (a) the \textit{degree to which social influence occurs} in a text; (b) one or more of 20 \textit{social influence techniques}; (c) the possible \textit{intentions} of the person exerting influence and the \textit{clarity} of those intentions; (d) possible \textit{consequences} of giving in to the influence and their \textit{severity}; (e) the \textit{reaction} of the influenced person, the level of \textit{resistance} against the influence, and whether they eventually \textit{submitted}; (f) the annotator's \textit{certainty} level; and (g) optional \textit{comments}.

\textit{Submission} was a binary decision (submitted/not submitted). \textit{Reactions}, \textit{intentions}, and \textit{consequences} (\textit{RICs}) were entered as free text, with each item recorded separately. The remaining questions had a 5-level scale, and all questions had a "hard to determine" option.

Every annotator annotated 204 or 205 texts in three rounds: first, they received a set of 30 texts (the \textit{Pre} set) for one week. Then, they annotated the main part of 174 or 175 texts for 2 weeks (the \textit{Main} set). Finally, they received the same 30 texts as at the beginning and labeled them from scratch (\textit{Post} set), again for one week. The \textit{Pre/Post} set of 30 texts is the main source for comparison. Before starting, the annotators participated in a technical training session on the online annotation tool (Argilla\footnote{\url{https://argilla.io}}), as well as a training session on social influence and the definitions used. After the first round of 30 texts, a calibration session was organized to answer questions and clarify technical misunderstandings. The session did not use any text as an example to ensure that it did not interfere with specific judgments. Our findings address both the \textit{Main} and \textit{Pre/Post} phases. Detailed annotator instructions, guidelines, answers from the calibration session, code, and definitions are available online\footnote{\url{https://github.com/MaciejMarkiewicz/annotator-competence-growth}}.

\subsection{Annotator demographics}

The annotation process was conducted by a group of 25 annotators, including 14 women and 11 men, with a mean age of 34 years. The annotator group was intentionally structured to represent diverse social and professional perspectives. It consisted of five educators (pedagogues), five psychologists, five adolescents, five parents of adolescents, and five communication experts.
Most had completed higher education (n = 19), five were students, and one had primary education. This diversity was intended to reduce perspective-related bias.

\section{Quantitative analysis methodology}\label{quantitative_methodology}

We assess annotator competence through data quality, defining increased competence as higher-quality work or equivalent quality in less time. We analyze quantitative differences in the identification of social influence between annotation rounds. Changes include technique labels, the number of "hard to determine" answers, reported answer certainty, and the number of free text \textit{RIC} responses. Since there are no gold standard labels and deriving them from the dataset itself would be biased (this is captured by agreement), we cannot treat an increase or decrease as desirable without qualitative confirmation. Specifically, we employed the following criteria: 

\subsection{Inter- and intra-annotator agreement} 
Agreement for technique identification was assessed using Krippendorff's alpha with Jaccard distance between sets, where each annotation was treated as a set of identified techniques. Inter-annotator agreement was calculated across all annotators and texts, while intra-annotator agreement measured the consistency between each annotator's \textit{Pre} and \textit{Post} labels for the same texts. Statistical significance was evaluated via bootstrap resampling of texts (10,000 iterations), preserving the dependency structure within the texts.

\subsection{Per-text annotation time change} 
The per-text time change was computed as the difference between consecutive submission timestamps for each annotator, with annotations ordered chronologically. To exclude outliers (very short times indicating skipped texts or very long times indicating breaks), a 1--20 minute window was applied. This analysis was conducted on both the \textit{Pre/Post} and \textit{Main} datasets. Statistical comparisons used Mann-Whitney U tests. Due to the nature of the observations, this method may not be fully accurate for precise time measurements but is relevant for capturing the trend.

\subsection{LLM training}
A direct measurement of data quality is the performance of AI models trained with it. To assess this, we evaluated the performance of LLMs in detecting social influence techniques by comparing the use of \textit{Pre} and \textit{Post} datasets for training and using the \textit{Main} dataset for testing. 

Social influence technique detection was formulated as a multi-label classification problem. We tested two aggregation methods to establish ground truth labels:
majority voting ($AC_{MV}$) and 2 annotator consensus ($AC_2$). Ground truth labels consisted of techniques identified by 3 or 2 out of 5 annotators, respectively. Some texts were labeled as lacking social influence, indicating that annotators did not detect any social influence in the text. 
For the sake of this experiment, instances where consensus could not be established were removed from the data. This approach reduced the \textit{Main} set from 871 to 869 ($AC_2$) or 800 ($AC_{MV}$) examples, and the \textit{Pre/Post} sets from 150 to 149 ($AC_2$) or 139 ($AC_{MV}$) examples. 

We evaluated both in-context learning (ICL) and supervised fine-tuning (SFT). ICL was conducted on \textit{DeepSeek-V3.2} (with the temperature set to 0) for $n \in \{0, 3, 10, 30\}$ shots, averaged over 10 runs with distinct seeds. SFT was performed on \textit{Llama-3.1-8B-Instruct} (batch size=1, LR=$1e^{-5}$, 2 epochs). Model performance was measured using Jaccard similarity.

\subsection{Annotators' self-perception of competence}

Two questionnaires captured changes in self-perceived competence. Along with the last one, after completing \textit{Post} annotations, the annotators were asked an additional open-ended question: \textit{"What did you learn by taking part in the study?"}.

\subsubsection{SIR-SC -- Social Influence Recognition Self-Competence}
questionnaire was developed, consisting of two sub-scales: \textit{Perceived Competence} (8 items on self-evaluated recognition skills) and \textit{Self-Confidence} (4 items on judgment certainty). Responses used a five-point Likert scale. Responses were rated on a five-point Likert scale (from 1 -- strongly disagree to 5 -- strongly agree). In both sub-scales, total scores were averaged. Higher scores indicate a higher level of perceived competence or self-confidence, respectively.

\subsubsection{NASA-TLX} 

\cite{hart1988development} was used to assess participants’ subjective workload associated with the annotation task. The instrument evaluates perceived workload across six dimensions: \textit{mental demand, physical demand, temporal demand, performance, effort,} and \textit{frustration}. On each dimension, participants provided a separate rating for a single, dedicated question on a 0–20 scale, reflecting their subjective assessment of the respective aspect of workload.
In the present study, the raw NASA-TLX score (Raw-TLX) was applied; no weighting procedure was used. The overall workload score was calculated by aggregating the ratings across the six dimensions. NASA-TLX was selected due to its established reliability and validity in assessing subjective workload in human-system interaction and task-based performance studies.

\section{Qualitative study methodology}\label{qualitative_methodology}

Two qualitative analyzes were performed to gain a deeper insight into the findings on the development of competences among annotators in identifying social influence.  
To this aim, two independent analyzes were conducted: a content analysis of annotators' free-text responses, followed by in-depth interviews.

\subsection{Content analysis of free-text responses}

For content analysis, we adopted a conventional content analysis methodology based on \cite{Hsieh2005}.  
The analysis of annotators' responses primarily focused on identifying differences in \textit{RICs} at the semantic level and recurring change patterns between two time points (\textit{Pre} and \textit{Post} annotations). 
Based on the data, we manually defined qualitative criteria for changes in \textit{RICs}. 

For all three, it included \textit{the number of semantically different observations}, a \textit{breadth of perspective} (narrowing, broadening, no change, reformulation), the \textit{level of detail in descriptions} (more, less, no change), the \textit{style of language} (formal, casual), and \textit{thematic categories} of items specific to each.

For the consequences of social influence, the following additional criteria were defined: \textit{reversibility} (more focus on reversible, more focus on irreversible, no change), \textit{inevitability} (more focus on potential, more focus on certain, no change), \textit{reach} (individual, group, social), and \textit{time horizon} (shorter, longer, no change). Thematic categories were \textit{psychological, social, economic, health, moral, and legal}.
For the intentions of a person exerting influence, the additional criteria included a change in \textit{focus concerning thematic categories}: intentions related to behaviors, actions, and facts, or related to motivations, goals, and mental processes. For the reactions of the influenced person, we also assessed the change in focus. The additional \textit{thematic categories} are \textit{emotional, cognitive, and behavioral}. Detailed definitions and codes are available in the supplementary material.

Next, following \cite{tai2024examination}, we performed LLM-assisted coding (using \textit{gpt-4o-mini}) of the above criteria, followed by expert verification. An LLM was only involved in the coding task and was not provided with information about the source (\textit{Pre} or \textit{Post} dataset) of each response. The final analysis and interpretation of the results were performed by the authors.

\subsection{A thematic analysis of in-depth interviews}

Next, a semi-structured interview was conducted after the annotation process with five annotators, one from each group. Semi-structured interviews allow for balancing comparability between participants with the flexibility to pursue novel themes that emerge during the conversation \cite{DeJonckheere2019}. The interview consisted of 13 open-ended questions to provide detailed information and reflections on the changes occurring in the process of double annotation of the same texts in terms of identifying and justifying social influence.

To analyze the interview responses, we adopted thematic analysis as described by Braun and Clarke \cite{BraunClarke2006}, which provides a systematic framework for identifying, analyzing, and reporting patterns within qualitative data. The responses of the respondents were recorded, transcribed, and analyzed in four described categories of the annotation process: (i) \textit{changes in the interpretation of texts}, (ii) \textit{changes in speed, efficiency, and confidence in recognized techniques}, (iii) \textit{changes in critical attitudes toward AI-generated texts}, and (iv) \textit{self-perceived ability to explain the social influence mechanisms to others}. In addition, we received information about annotators' strategies for memorizing techniques and difficulties in the annotation process. 

This framework supports the transparency and credibility of the findings derived from semi-structured interview transcripts.
We applied a hybrid inductive-deductive thematic analysis. The initial codes were grounded in the interview guide, which led to the emergence of thematic categories. At least two researchers-experts independently conducted the coding, and disagreements were resolved through discussion. Finally, the themes were organized into higher-level categories that reflected the growth in competence.

\section{Results and analyses}\label{results}

A high number of annotation changes was encountered in the \textit{Pre/Post} data. Table~\ref{tab:pre_post_groups_delta} presents the changes. Expert groups showed significant increases (Mann-Whitney U test, $p<0.05$ for reactions, $p<0.001$ for intentions and consequences), while non-experts did not. 

\begin{table}[htbp]
\caption{\textit{Pre}, \textit{Post}, and change ($\Delta$ = Post--Pre) in counts of techniques, reactions, intentions, and consequences by group. The biggest absolute $\Delta$ is marked in bold.}
\label{tab:pre_post_groups_delta}
\centering

\setlength{\tabcolsep}{4.3pt}
\scriptsize 
\begin{tabular}{lrrrrrrrrrrrr}
\toprule
\multirow{2}{*}{\textbf{Group}}
& \multicolumn{3}{c}{Techniques}
& \multicolumn{3}{c}{Reactions}
& \multicolumn{3}{c}{Intentions}
& \multicolumn{3}{c}{Consequences} \\
\cmidrule(lr){2-4}
\cmidrule(lr){5-7}
\cmidrule(lr){8-10}
\cmidrule(lr){11-13}
& Pre & Post & $\Delta$
& Pre & Post & $\Delta$
& Pre & Post & $\Delta$
& Pre & Post & $\Delta$ \\
\midrule
Expert      
& 2.06 & 2.29 & \textbf{0.23}
& 1.40 & 1.83 & \textbf{0.43}
& 1.49 & 1.91 & \textbf{0.42}
& 2.28 & 3.27 & \textbf{0.99} \\
Non-expert 
& 2.68 & 2.51 & -0.17
& 1.42 & 1.42 & 0.00
& 1.69 & 1.62 & -0.07
& 2.45 & 2.51 & 0.06 \\
Overall     
& 2.43 & 2.42 & -0.01
& 1.41 & 1.58 & 0.17
& 1.61 & 1.73 & 0.12
& 2.38 & 2.81 & 0.43 \\
\bottomrule
\end{tabular}
\end{table}

Qualitative analysis pointed to differences in how the annotators articulated social influence qualities. \textit{Post} responses were generally more detailed and offered a broader perspective, as presented in Table~\ref{tab:qualitative_changes}. This corresponds well with an observed increase in the number of \textit{RICs}, suggesting that new elements contain novel observations. This is further evaluated by assessing semantically different concepts and their corresponding \textit{thematic categories}. Unlike quantitative analyzes, this effect is observed for all groups, not only experts, but it is more prevalent among them (particularly among the \textit{communication experts} group).

\begin{table}[htbp]
\caption{Distribution of categorical changes between \textit{Pre} and \textit{Post} annotations across consequences, reactions, and intentions.}
\label{tab:qualitative_changes}
\centering
\setlength{\tabcolsep}{4pt}
\scriptsize 
\begin{tabular}{lrrr}
\toprule
\textbf{Change Type} & \textbf{Consequences} & \textbf{Reactions} & \textbf{Intentions} \\
\midrule
Breadth: Broadening & \textbf{275 (44.2\%)} & \textbf{230 (46.7\%)} & \textbf{261 (39.7\%)} \\
Breadth: Narrowing & 142 (22.8\%) & 103 (20.9\%) & 183 (27.9\%) \\
Breadth: No change & 196 (31.5\%) & 159 (32.3\%) & 199 (30.3\%) \\
\midrule
Detail: More details & \textbf{341 (54.8\%)} & \textbf{256 (52.0\%)} & \textbf{395 (60.1\%)} \\
Detail: Less details & 201 (32.3\%) & 153 (31.1\%) & 231 (35.2\%) \\
Detail: No change & 80 (12.9\%) & 83 (16.9\%) & 31 (04.7\%) \\
\bottomrule
\end{tabular}
\end{table}

\subsection{Changes in the formulation of free-text responses}

\begin{figure}[htbp]
    \centering
    \includegraphics[width=1\linewidth]{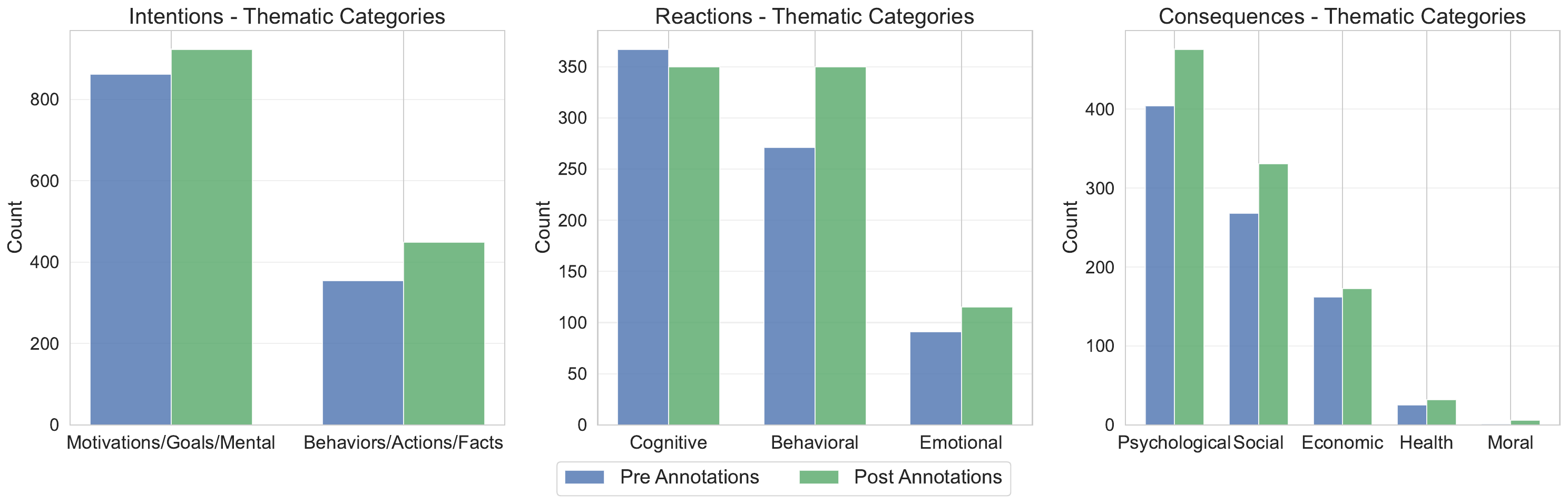}
    \caption{Shifts in the number of individual \textit{RICs} by \textit{thematic categories}.}
    \label{fig:thematic_categories}
\end{figure}

Figure~\ref{fig:thematic_categories} shows thematic category distributions across \textit{RICs}. Post annotations exhibited more balanced distributions across categories, consistent with increased thematic coverage.
Initial \textit{Pre} annotations described the intentions behind social influence mainly as \textit{motivations or goals} rather than \textit{behaviors or actions} ($861$ vs $354$ distinct intentions). \textit{Post} responses added slightly more of the latter ($923$ vs $449$), resulting in a more balanced distribution. Analyzes of language change indicate the adoption of  a more formal ($194$ shifts from casual to formal, $74$ vice versa) language, which may be a sign of greater fluency in describing the intentions.

For reactions to social influence, a similar change in the balance of the distribution of \textit{thematic categories} may be observed, with an increase in the least common categories. Language analyzes support this, with a more common use of casual language ($154$ changes from formal to casual, $74$ vice versa), which may be associated with a higher number of concrete \textit{behavioral} reactions.

In describing the consequences of social influence, all participant groups tended to preserve a similar temporal horizon, with no marked shifts in the perceived time span of effects. 
A slight shift towards more formal language ($118$ vs $35$ changes) was observed. In terms of \textit{thematic categories}, a slight increase in the least common categories was observed, but there was no significant change in the overall distribution. 
A more noticeable change emerged in the dimensions of \textit{inevitability} and \textit{range}. \textit{Post} annotations more often framed the consequences of social influence as potential, hypothetical, or conditional rather than as definite outcomes ($335$ changes towards potential, $74$ towards definite, $223$ no change), indicating a move toward greater interpretative caution and openness in consequence formulation. The perception of \textit{range} has broadened, with a shift from noticing mostly consequences concerning one person ($163$ individual, $43$ group, $1$ systematic in \textit{Pre}) to more people ($604$ individual, $198$ group, $7$ systematic in \textit{Post}). This indicates a higher awareness of the outcomes.

In general, qualitative analyzes suggest that changes usually involve the enrichment, refinement, or reconsideration of the emphasis between categorical components, rather than a significant conceptual change. These observations may indicate the increased competence of the annotators.

\subsection{Annotation Speed and Efficiency} 

Analysis of annotation time revealed a significant decrease between the first and the repeated rounds. In the first round, annotators spent an average of 6.56 minutes per text (median = 5.49, SD = 3.99), while in the repeated round, it decreased to an average of 5.86 minutes (median = 4.69, SD = 3.74). This difference was statistically significant according to the Mann-Whitney U test ($U = 171438.0$, $p = 0.0009$).

\begin{figure}[htbp]
    \centering
    \includegraphics[width=1\linewidth]{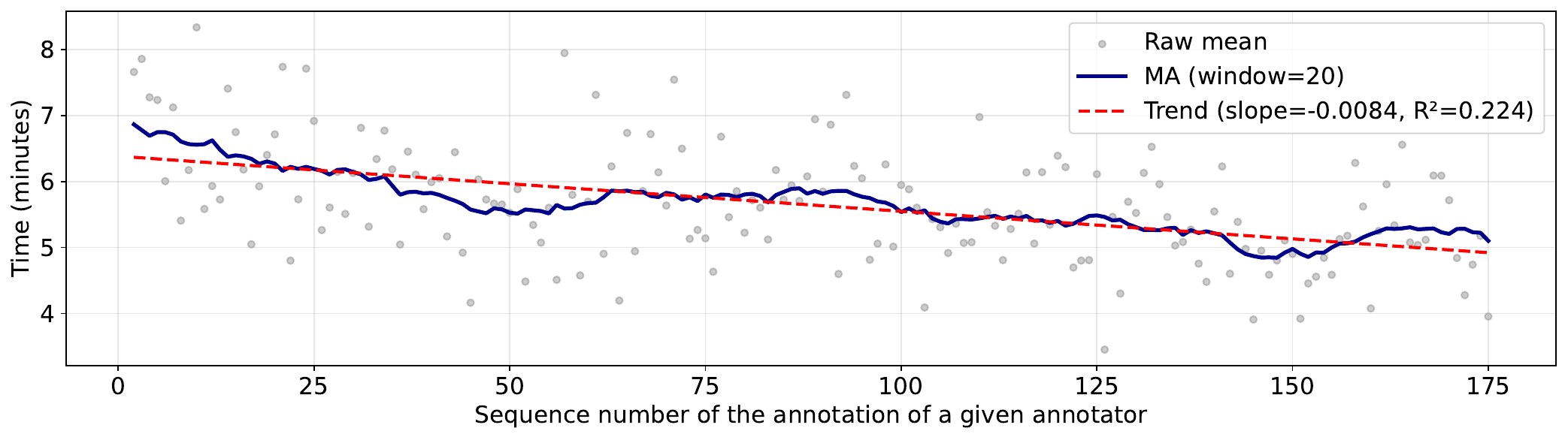}
    \caption{Annotation time change over the course of the process.}
    \label{fig:time}
\end{figure}

Trend analysis across the entirety of the \textit{Main} annotation revealed a consistent pattern of decreasing annotation time. The overall combined trend showed a significant negative slope ($\beta = -0.0084$, $R^2 = 0.224$, $p < 0.001$, Figure~\ref{fig:time}), indicating that annotators became progressively faster throughout the process.

\subsection{Annotator agreement}

For technique classification, inter-annotator agreement (Krippendorff's $\alpha$ with Jaccard distance) improved slightly from $\alpha = 0.319$ to $\alpha = 0.327$, though this difference was not statistically significant. Intra-annotator agreement, measured as the pairwise $\alpha$ between each annotator's \textit{Pre} and \textit{Post} labels, was higher than inter-annotator agreement (mean $\alpha$ = 0.442, $SD$ = 0.121), which is consistent with the characteristics of subjective tasks but is still quite low, suggesting difficulty \cite{abercrombie-etal-2025-consistency}. However, perfect intra-annotator agreement would exclude the possibility of competence increase.

\subsubsection{Group-Level Differences} 

Expert annotators demonstrated both higher inter- ($\alpha = 0.383 \rightarrow 0.405$) and intra-annotator agreement (mean $\alpha = 0.514$, $SD = 0.069$) compared to non-experts (inter- $\alpha = 0.290 \rightarrow 0.286$; intra- mean $\alpha = 0.394$, $SD = 0.124$). Notably, inter-annotator agreement improved for the expert group ($+0.023$) while remaining essentially unchanged for non-experts ($-0.004$), suggesting that the annotation process might have reinforced convergence primarily among those with prior domain knowledge.

\subsection{Technique Stability} 

The analysis revealed systematic shifts in annotation patterns. While the total number of technique labels remained stable (1821 vs 1816), specific assignments changed substantially, with an average retention rate of 65\% across techniques. The most consistent techniques were those with clear behavioral markers: \textit{Door-in-the-face} and \textit{Show disappointment} (75\% retention), \textit{Flattery} (74.5\%), and \textit{Give to take} (71.2\%). In contrast, abstract affective techniques showed lower stability, with \textit{Liking} retained only 40.8\% of the time.

The most frequent substitution patterns -- \textit{Liking} $\rightarrow$ \textit{Labeling} (15 instances), \textit{We are exceptional} $\rightarrow$ \textit{The ``We'' rule} (11), and \textit{Gratitude} $\rightarrow$ \textit{Give to take} (10) -- suggest that annotators refined broad, intuitive categorizations into more specific, behaviorally-defined techniques. This is reflected at the aggregate level: \textit{Labeling} showed the largest net increase (+46.4\%), while \textit{Liking} showed the largest decrease ($-$25.2\%). These patterns indicate that the annotation process might have enhanced annotators' ability to distinguish between conceptually overlapping techniques.
 
\subsection{LLM training}

\textit{Post} data consistently improved model performance across all settings (Table \ref{tab:training}. For ICL, performance scaled positively with the number of few-shot examples ($n$), and the \textit{Pre/Post} gap widened as $n$ increased. All ICL results were above a zero-shot baseline of $0.3833$. While the difference was negligible at $n=3$ ($\Delta=+0.0011$), the \textit{Post} set demonstrated a distinct advantage at $n=30$ ($\Delta=+0.0149$). 
For SFT, \textit{Llama-3.1-8B-Instruct} improved from an untrained baseline of $0.1235$. \textit{Post} data yielded a small but positive delta of ($\Delta=+0.0069$), consistent with the trajectory observed in ICL experiments. The size of the improvement is limited due to a small number of training samples (149). 

All presented data used the $AC_{MV}$ consensus. We found that using the looser $AC_2$ consensus yielded a much worse overall performance, with insignificant differences observed between conditions ($\Delta < SD$).

\begin{table}[htbp]
\caption{Model performance difference when using \textit{Pre} and \textit{Post} data for training.}
\label{tab:training}
\centering
\scriptsize
\setlength{\tabcolsep}{4pt}
\begin{tabular}{lrrr}
\toprule
\textbf{Model and setting} &
\textbf{\textit{Pre} (Jaccard$\uparrow$)} &
\textbf{\textit{Post} (Jaccard$\uparrow$)} &
\textbf{$\Delta$ Jaccard} \\
\midrule
DeepSeek-V3.2 (3-shot)  & 0.4523 ($\pm$ 0.0058) & \textbf{0.4534} ($\pm$ 0.0058) & +0.0011 \\
DeepSeek-V3.2 (10-shot) & 0.4874 ($\pm$ 0.0044) & \textbf{0.4962} ($\pm$ 0.0014) & +0.0088 \\
DeepSeek-V3.2 (30-shot) & 0.5214 ($\pm$ 0.0048) & \textbf{0.5363} ($\pm$ 0.0033) & +0.0149 \\
\midrule
Llama-3.1-8B-Instruct (SFT) & 0.2684 & \textbf{0.2753} & +0.0069 \\
\bottomrule
\end{tabular}
\end{table}

\subsection{Workload perception}
The analysis of NASA-TLX scores revealed statistically significant differences between the measurements. The mean TLX score increased from 50.80 (SD = 15.95) after \textit{Pre} annotations to 60.56 (SD = 19.21) after \textit{Post} annotations.

\begin{figure}[htbp]
    \centering
    \includegraphics[width=1\linewidth]{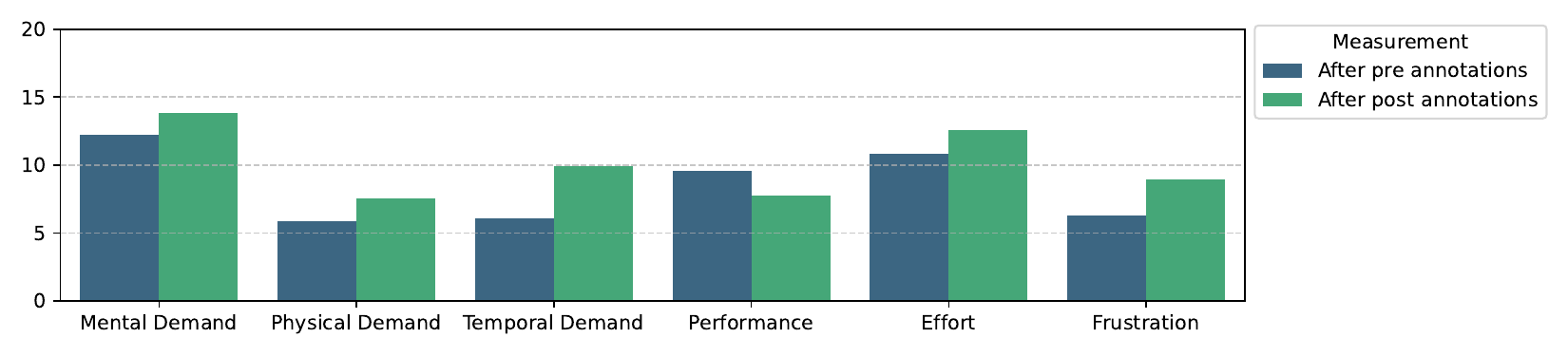}
    \caption{Comparison of mean scores of NASA-TLX dimensions between measurements.}
    \label{fig:tlx}
\end{figure}

At the dimensional level, the second measurement showed higher scores for mental demand, physical demand, temporal demand, effort, and frustration. Conversely, the performance score significantly decreased, indicating a lower subjective assessment of task performance despite a higher workload (see \ref{fig:tlx}). According to the literature, this is a characteristic of the learning phase or "conscious incompetence", where gaining knowledge makes learners more aware of their own mistakes \cite{Mohamed2014}.

\subsection{Annotators' perception of competence development and confidence}

\begin{figure}[htbp]
    \centering
    \includegraphics[width=0.9\linewidth]{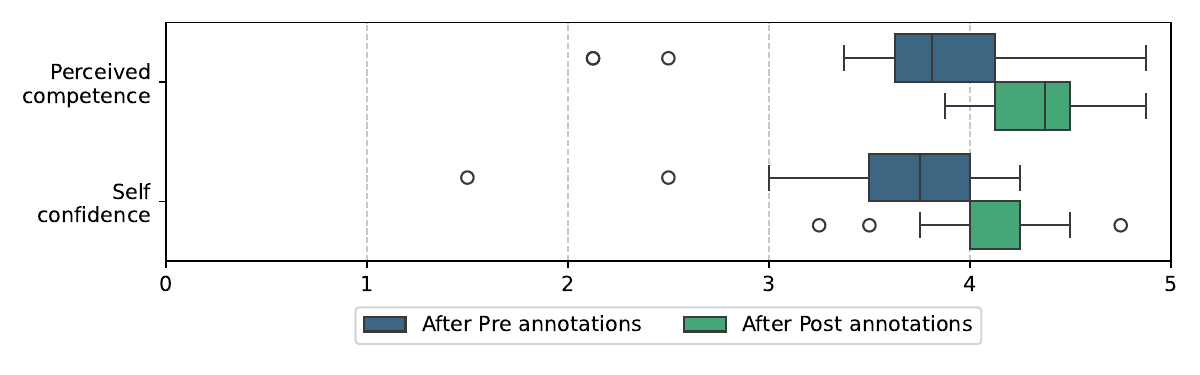}
    \caption{Box Plots of \textit{Perceived Competence} and \textit{Self-confidence} across measurements.}
    \label{fig:sirsc}
\end{figure}

Figure~\ref{fig:sirsc} presents the \textit{Perceived Competence} and \textit{Self-Confidence} \textit{SIR-SC} scales at both measurement points. \textit{Perceived Competence} showed a notable increase with a large effect size  (Cohen’s $d = 0.567$), indicating a substantial impact on participants' self-assessed mastery. \textit{Self-Confidence} also improved significantly, with a medium effect size (Cohen’s $d = 0.388$).

Survey results aligned with answer certainty from annotation data: certainty scores increased significantly across all groups ($p < 0.001$, Mann-Whitney U test) from $M = 3.67, SD = 0.73$ to $M = 3.92, SD = 0.70$.

Qualitative analysis of open responses revealed that the most frequently mentioned gains related to distinguishing social influence (31.7\%), followed by technique detection, increased awareness, and the ability to name specific mechanisms (each 19.5\%). Less common were awareness of consequences (7.3\%) and knowledge of assertive responses (2.4\%). These answers are reflected and elaborated upon in the analysis of extended interviews:

\textbf{Changes in the interpretation of texts} were described as the reconstruction of additional context, re-interpretation of irony as manipulation, and reassessment of previously neutral statements for hidden intent and consequences. One annotator framed this as a deepened analysis rather than a revision. The main challenge was differentiating conceptually similar techniques, especially those tied to group identity, and deciding when multiple techniques were applied to one text. With continued annotation, the boundaries became clearer. These answers suggest an increase in sensitivity to subtle cues. However, some changes between rounds were not fully understood, despite being noticed.

\textbf{Development of speed, confidence, and detection ability} over time was reported by most annotators, consistent with previous findings. They noticed their judgments shifting from moderate to more polarized high-confidence ratings, reporting that competence developed “naturally” through continual reference to definitions and examples. One interviewee diverged, stating that later texts felt more ambiguous and cognitively demanding, leading to longer deliberations and more frequent uncertainty.

\textbf{Increasing critical awareness of AI-generated content} was mentioned by all participants, especially regarding AI-generated texts. Texts perceived as unrealistic elicited negative emotions. One interviewee noted a pre-existing critical stance due to professional experience but still reported increased awareness of the scale and pervasiveness of influential content. 
 
\textbf{The ability to explain manipulation to others} has improved, as reported by all participants. Getting familiar with formal terminology was repeatedly described as enabling clearer explanations of previously intuitive impressions, thereby increasing confidence and ease of communication. Some had already discussed techniques with their children or peers. Several also noted that annotation is likely one component of greater education to counteract social influence, which directly references the scope of this study.
 
\textbf{Additional remarks} included developing deliberate learning strategies described by four participants and the experienced spontaneous detection of manipulation in everyday communication. Annotators relied mainly on repeated exposure to guidelines and extensive practice. Prior domain familiarity (in expert groups) supported faster learning, while unfamiliar techniques required more intentional practice and consultation of training materials. One participant noted recurring lexical/situational patterns (e.g., repeated brand-name cues) that facilitated classification, while others did not report this.

\section{Discussion}

Annotation of social influence is difficult and highly subjective, as reflected by moderate inter- and intra-annotator agreement \cite{abercrombie-etal-2025-consistency}. This is further supported by the fact that both measures are higher in the expert group compared to non-experts. Although changes between annotation rounds are necessary to observe improvement, groups with higher intra-annotator agreement (fewer changes) tended to exhibit more characteristics of an increase in data quality.

All annotators learned to perform the task more quickly, and their self-perceived competence and confidence increased significantly. Annotators also began to notice the effects of learning about social influence extending beyond the scope of their work. An improved perception of the \textit{intentions} of a person exerting social influence, its possible \textit{consequences}, and a greater awareness of the \textit{reactions} of the target are supported by an increase in the mean number of these observations. The better quality of these was confirmed in a qualitative analysis for all groups, with a broadened perspective, a higher level of detail, and wider thematic coverage. 

Finally, our intention was for annotators to be active participants in designing the system aimed at supporting adolescents in resisting social influence. Therefore, high annotator competence and their involvement in defining educational goals and processes, further implemented with the use of LLMs trained on the annotated data, are essential.

\section{Conclusions}

Regarding RQ2 and RQ3, improved model performance and increased annotator agreement suggest higher data quality over time.
In response to RQ1, the observed effects may indicate that annotator competence increases over the course of the annotation process, particularly among annotators who are already experts. It is therefore necessary to monitor annotators' competence development, as it has a direct impact on the performance of models trained on their data.

The process of annotating social influence can be seen as an effective educational intervention. It strengthens awareness of persuasive intent, improves recognition of subtle manipulation, and increases participants’ confidence in communicating about manipulation with others. Our findings support the integration of annotation-based activities into educational programs addressing AI literacy and protecting youth from harmful, persuasive content.

Directions for future work in this area include investigating larger participant groups across multiple subjective tasks. The additional incorporation of non-subjective tasks could enable external, quantitative competence assessment and provide a more direct interpretation of agreement measures.

\begin{credits}
\subsubsection{\ackname} 

This work was financed by 
(1) the National Science Centre, Poland, project no. 2021/41/B/ST6/04471;
(2) the statutory funds of the Department of Artificial Intelligence, Wrocław University of Science and Technology;
(3) the Polish Ministry of Education and Science within the programme “International Projects Co-Funded”;
(4) the European Union under the Horizon Europe, grant no. 101086321 (OMINO). However, the views and opinions expressed are those of the author(s) only and do not necessarily reflect those of the European Union or the European Research Executive Agency. Neither the European Union nor European Research Executive Agency can be held responsible for them.
\end{credits}

\bibliographystyle{splncs04}
\bibliography{bibliography}

\end{document}